\documentclass[11pt]{article}

\usepackage{times}
\usepackage{amsmath}

\DeclareMathOperator*{\argmin}{arg\,min}

\usepackage[numbers,comma]{natbib}

\usepackage{amssymb,latexsym,amsfonts,relsize}
\usepackage{graphicx}
\usepackage{fancybox}
\usepackage{xcolor}
\usepackage{mathtools}
\usepackage[utf8]{inputenc}
\usepackage[english]{babel}
\usepackage[T1]{fontenc}
\usepackage{helvet}
\usepackage{tikz}

\newcommand{\edo}{\end{document}}

\newcommand{\pic}[2]{\includegraphics[scale=#1]{#2}}
\newcommand{\picc}[2]{\begin{center}\pic{#1}{#2}\end{center}}
\newcommand{\minp}[2]{\begin{minipage}{#1\textwidth}#2\end{minipage}}
\newcommand{\comment}[1]{}
\newcommand{\ms}{\medskip}
\newcommand{\itt}{$\bullet\;$} 

\newcommand{\red}[1]{{\textcolor{red}{#1}}}
\newcommand{\green}[1]{{\textcolor{green}{#1}}}

\newcommand{\blue}[1]{{\textcolor{blue}{#1}}}
\newcommand{\dgreen}[1]{{\textcolor{darkgreen}{#1}}}
\definecolor{darkgreen}{RGB}{51,102,0}           

\newcommand{\beqn}{\begin{eqnarray*}}
\newcommand{\eeqn}{\end{eqnarray*}}
\newcommand{\bi}{\begin{itemize}}
\newcommand{\ei}{\end{itemize}}
\newcommand{\ben}{\begin{enumerate}}
\newcommand{\een}{\end{enumerate}}
\newcommand{\norm}[1]{\left\Vert #1 \right\Vert}

\newcommand{\R}{{\mathbb R}}  

\newcommand{\negskipsmall}{}

\newcommand{\newsl}[1]{\subsubsection*{#1}}

\newcommand{\X}{{\mathbb X}}

\usepackage[export]{adjustbox}

\newcommand{\gV}{\nabla \V}
\newcommand{\V}{{\cal L}}
\newcommand{\kl}{{\mathcal K}{\mathcal L}}
\newcommand{\ki}{{\mathcal K_\infty}}
\newcommand{\kk}{{\mathcal K}}
\newcommand{\abs}[1]{\left\vert #1 \right\vert}

\newcommand{\xtou}{x(t,\xo,u)}
\newcommand{\xsou}{x(s,\xo,u)}  

\newcommand{\xo}{x_{\scriptscriptstyle 0}}
\newcommand{\normi}[1]{\norm{#1}_{\infty}}

\newcommand{\ggt}{\gamma}

\newcommand{\Rn}{\R^{n}}

\newcommand{\Ex}{\mathbb{E}}

\newcommand{\dX}{\partial\X}

\newcommand{\stt}{\,|\,} 

\newcommand{\target}{\mathcal{T_{\V}}}
\newcommand{\crit}{\mathcal{C}_{\V}}
\newcommand{\st}{\,|\,}

\newcommand{\OptSlideKoopman}[1]{}   

\newcommand{\dist}{\mbox{dist}\,}

\tikzstyle{neuron} = [draw, fill=white, circle, 
    minimum height=3.5em, minimum width=1em, style=thick]
\newcommand{\Kdom}{\mathcal{K}}

\newcommand{\gray}[1]{{\color{gray}{#1}}}
\renewcommand{\green}[1]{\dgreen{#1}} 
\newcommand{\brown}[1]{{\color{brown}{#1}}}

\newcommand{\redbox}[1]{{\color{red}{\boxed{\color{black}{#1}}}}}
\newcommand{\itb}{$\bullet\;$}
\newcommand{\mymatrix}[1]{\begin{array}{cccccccccccc}#1\end{array}}
\newcommand{\argL}{k}

\renewcommand{\X}{\Dom}

\newcommand{\Dom}{D}
\newcommand{\loss}{{\mathcal L}}
\newcommand{\losst}{\ell}
\newcommand{\gradloss}{\nabla \loss}
\newcommand{\lossmin}{\underline{\loss}}

\usepackage{xspace}
\newcommand{\PLI}{P\L{}I\xspace}
  \newcommand{\gPLI}{\mbox{\textit{gl}-\PLI}\xspace}
 \newcommand{\sgPLI}{\mbox{\textit{sgl}-\PLI}\xspace}
\newcommand{\satPLI}{\mbox{\textit{sat}-\PLI}\xspace}
\newcommand{\lPLI}{\mbox{\textit{loc}-\PLI}\xspace}

\newcommand{\twoif}[4]{
\left\{ \begin{array}{ll}#1&#2\\#3&#4\end{array}\right.
}
\newcommand{\kmin}{\underline{k}}
\newcommand{\kfeedback}{k}
\newcommand{\compfun}{\alpha}
\newcommand{\timearg}{} 

\newcommand{\kpl}{\gPLI}
\newcommand{\kplloc}{\lPLI}
\newcommand{\ks}{\satPLI}

\newcommand{\pd}{{\mathcal P}{\mathcal{D}}}

\newcommand{\KK}{k}
\newcommand{\K}{\mathbf{k}}
\newcommand{\calC}{\mathcal{C}}
\newcommand{\ksetmin}{\underline{K}}

\newcommand{\strict}{{\mathcal S}_{\loss}}

\title{\mbox{Some remarks on gradient dominance and LQR policy optimization
\footnote{This is a short paper summarizing 
the slides presented at my keynote
at the 2025 L4DC (Learning for Dynamics \& Control Conference) in Ann
Arbor, Michigan, 05 June 2025. A partial bibliography has been added.
}}}

\author{Eduardo Sontag\footnote{Supported in part by grants
AFOSR FA9550-2110289 \& 22RT0159, ONR N00014-21-1-2431, NSF/DMS-2052455.}
\\
Northeastern University\\
Electrical and Computer Engineering \& BioEngineering Departments}

\begin{document}

\def\today{}

\maketitle

\begin{abstract}

Solutions of optimization problems, including policy optimization in
reinforcement learning, typically rely upon some variant of gradient
descent. There has been much recent work in the machine learning,
control, and optimization communities applying the Polyak-Łojasiewicz
Inequality ({\PLI}) to such problems in order to establish an exponential
rate of convergence  (a.k.a. ``linear convergence'' in the
local-iteration language of numerical analysis) of loss functions to
their minima under the gradient flow. Often, as is the case of policy
iteration for the continuous-time LQR problem, this rate vanishes for
large initial conditions, resulting in a mixed globally linear /
locally exponential behavior. This is in sharp contrast with the
discrete-time LQR problem, where there is global exponential
convergence. That gap between CT and DT behaviors motivates the search
for various generalized {\PLI}-like conditions, and this paper addresses
that topic. Moreover, these generalizations are key to
understanding the transient and asymptotic effects of errors in the
estimation of the gradient, errors which might arise from adversarial
attacks, wrong evaluation by an oracle, early stopping of a
simulation, inaccurate and very approximate digital twins, stochastic
computations (algorithm ``reproducibility''), or learning by sampling
from limited data. We describe an ``input to state stability''
(ISS) analysis of this issue. We also discuss convergence and
{\PLI}-like properties of ``linear feedforward neural networks'' in
feedback control.
Much of the work described here was done in collaboration with
Arthur Castello B. de Oliveira, Leilei Cui, Zhong-Ping Jiang, and Milad Siami.

\end{abstract}

\section{Loss functions and gradient flows}


\negskipsmall

Suppose that we want to minimize a ``loss'' or ``cost'' function on an open subset $\Dom \subseteq  \Rn$:
\[
\loss \,:\; \Dom \rightarrow  \R \,:\;\; k \mapsto  \loss(k)
\]
(with no convexity assumptions).
We will assume that a minimum exists, and let:
\[
\ksetmin := \argmin_{k\in \Dom} \loss(k)\,,\;\;
\; \lossmin := \min_{k\in \Dom} \loss(k)\,.
\]
Often this minimum is achieved at just one point, $\ksetmin =\{\kmin\}$,
and this happens for example in the LQR application described next,
but we need not assume uniqueness at this point.

\textit{Example:} Consider the continuous-time LQR problem for a
simple one-dimensional integrator $\dot x = u$, $x(0)=1$.
Here $n\! =\! m\! =\! 1$, and we take the costs matrices as $1$.
The general problem is stated more precisely and generally later on).
The loss function is:
\[
\loss(k) \,:=\; \int_0^\infty  x(s)^2 + u(s)^2\,ds, \;\; u(s)=-kx(s)
\,,\;\; k\in \Dom = (0,+\infty )
\]
(i.e., $k$ stabilizes the closed loop system $\dot x = -kx$).
Evaluating the integral, we have that:
\[
  \loss(k) \;=\;  \frac{1+k^2}{2k} \,.
\]
Note that in this example, $\kmin=1, \, \lossmin=1$.
We will come back to this very special example later, to illustrate various results.
Gradient flows for LQR arise in \textit{policy optimization}
in reinforcement learning ``model-free'' searches for optimizing
feedbacks, as opposed to a value function model-based (Riccati) approach.


\newsl{Gradient flows}

The standard gradient flow for a (continuously differentiable) loss
$\loss : \Dom \rightarrow  \R$ is a set of ODEs for the components of $k$:
\[
{\dot k(t) \;=\; - \eta\,\gradloss(k(t))^\top\,, \;\;
k(t)\in \Dom\,, \; t\geq 0} \;\;  
\]
(from now on, for notational simplicity we'll set the ``learning rate'' $\eta$ to 1).
We view the gradient $\gradloss=$ as a row vector, hence the transpose.
One could of course study similar problems on Riemannian manifolds $M$,
where the gradient vector field of $\loss$ is defined by the duality condition
$g_p(\gradloss(p), X) = d\loss_p(X_p)$, where $d\loss_p$ is the exterior
derivative (or differential) of $\loss$ at $p$ and $g$ denotes the inner product
in the tangent space $T_pM$ at a point $p\in M$.
One can, and does, study similar problems for quasi-Newton, proximal
gradient, and other optimization flows.
The main motivation for the study of gradient flows is that
the classical \textit{steepest descent} iteration used when minimizing
the function $\loss$ is simply the Euler discretization of the gradient flow:
\[
k(t+h) = k(t) - h\,\gradloss(k(t))^\top
\]
where $h>0$ is the step size. This is a
discrete-time system associated to the continuous-time ODE.

We now turn to the study of convergence
$k(t)\rightarrow \ksetmin$, or (easier as a first step) convergence of the loss
$\loss(k(t))\rightarrow \lossmin$, as $t\rightarrow \infty$.

\newsl{Convergence}


\newcommand{\hessian}{{\mathcal H}(k)}

Let us define the
\textit{target}, \textit{critical}, and \textit{strict saddle} sets for the gradient
flows as follows:
\beqn
{\target} &:=& \left\{k \st {\cal L}(k)=\lossmin\right\}\\
{\crit}    &:=& \left\{k \st \gV(k)=0\right\}\\
\strict        &:=& \left\{k \st \gV(k)=0 \mbox{ and } \hessian \mbox{ has a negative eigenvalue}\right\}
\eeqn
where $\hessian$ is the Hessian of $\loss$ and as earlier $\lossmin= \min_{k\in \Dom} \loss(k)$.
The equilibria of the gradient flow are the critical points, and
$\target = \ksetmin$ is the set of points at which the loss is
globally minimized. 
The set $\strict$ consists of \textit{strict saddles}\footnote{The
terminology might be a little confusing, as a local maximum can be a
``saddle'' in this sense.} of $\loss$,
meaning those equilibria at which the linearization of
$\dot k(t) = -\gradloss(k(t))^\top$ is
exponentially unstable (has an eigenvalue with strictly
positive real part). Since the linearization of $\loss$ at a point $k$
is the symmetric matrix $\hessian$, the condition is that $\hessian$
should have a negative eigenvalue. 
Note that:
\[
{\target} 
\;\subseteq\; {\crit}
\,,\;\;
\strict
\;\subseteq\;{\crit}
\,,
\;\;\mbox{and}\;\;
\target \cap \strict = \emptyset\,.
\]
Now let us consider the relative loss (``regret'') along solutions of the gradient system:
\[
\losst(t):=\loss(k(t))-\lossmin \,.
\]
Clearly,
\[
\dot \losst(t) \;=\;
\gradloss(k(t)) \, \dot k(t) \;=\;
\gradloss(k(t)) \, [ - \gradloss(k(t))^\top ] \;=\;
-\norm{\gradloss(k(t))}^2 \;\leq\; 0\,.
\]
From this, we can conclude as follows:
\ben
\item
\textit{Precompact trajectories $k(t)$ approach $\crit$.}
\item
\textit{If ${\cal L}$ is a real-analytic function, then all
$\omega $-limit sets ($\subseteq\crit$) are \textit{single equilibria}.}
\item
\textit{Generically, precompact trajectories converge to $\crit\setminus\strict$.}
\een
Conclusion 1 follows immediately from the Krasovskii-LaSalle invariance principle. 
Conclusion 2 is \L{}ojasiewicz' Gradient Theorem, and in fact holds as
well for arbitrary loss functions (smooth but not necessarily
analytic) that satisfy the $\gPLI$ property to be described later.
Conclusion 3 is also valid for general nonlinear systems such as the
``overparametrized''
problem also discussed in the lecture, if there are non-isolated saddles,
including a continuous set of saddles.
This fact follows from the Center-Stable Manifold Theorem together
with a topological argument. The proof is given in
\cite{24cdc_arthur_nn_gradient},
and it is based on tightening
arguments from an analogous but slightly easier result for discrete
time systems \cite{2017_panageas_et_al_gradient_descent}.
We summarize as follows:
\ms

\minp{0.63}{%
\textbf{Theorem:}
Suppose that:
\bi
\item
$\V$ is real-analytic,
\item
$\crit = \target \sqcup \strict$
(i.e., $\crit\!\setminus\!\target \subseteq\strict$), and
\item
  all trajectories of $\dot k(t)=- \gradloss(k(t))^\top$
  are precompact.
\ei
}%
\minp{0.35}{\pic{0.25}{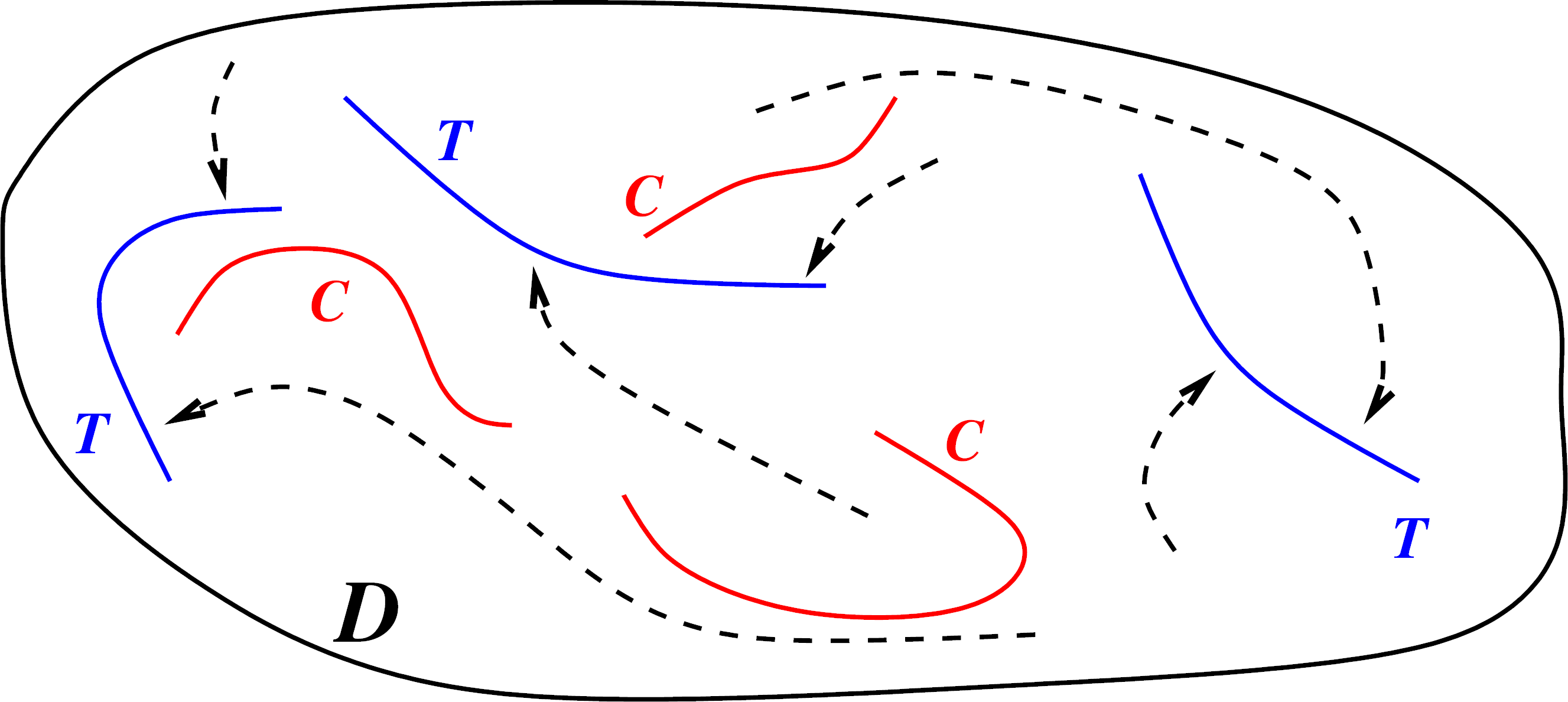}}

\ms
Then, except a set of measure zero, trajectories converge to \textit{points} in $\target$.

\ms\ms\ms

%
%
%
%

\section{Polyak-\L{}ojasiewicz Inequality (\PLI) and several variants}

Suppose now that $\loss$ satisfies the 
\textit{global Polyak-\L{}ojasiewicz Inequality}
gradient dominance condition:
\[
\redbox{\exists \,\lambda >0 \;\; \mbox{s.t.} \;\;\norm{\gradloss(k)} \geq \sqrt{
    \lambda  (\loss(k) - \lossmin)} \;\; \forall \, k \in D} \quad \red{(\gPLI)}
\]
where the estimate can be rewritten as
\[
\norm{\gradloss(k)}^2 \;\geq\;  \lambda  (\loss(k) - \lossmin)\,. 
\]
It can be shown that all strictly convex functions satisfy $\gPLI$, but the
converse is not true in general, i.e.\ there are many nonconvex
functions that satisfy a $\gPLI$ condition, as well as
similar but weaker
conditions, as we will remark soon for the LQR problem.
The main reason to look at $\gPLI$ is as follows.
Consider any solution of $\dot k(t) = - \gradloss(k(t))^\top$ and
recall that $\losst(t)=\loss(k(t))-\lossmin$.
One concludes \textit{global exponential convergence} of the relative loss to zero:
\[
\dot \losst(t) \leq  -\norm{\gradloss(k(t))}^2 \leq
-\lambda (\loss(k(t)) - \lossmin)
=
-\lambda \losst(t)
\;\Rightarrow \;
\; {\losst(t) \leq  e^{-\lambda t} \losst(0)}\,.
\]
This is all quite standard by now. A central point
of this talk is that \textit{often one does not have \textit{(global)} \PLI},
yet there are weaker versions that do hold.

%



\newsl{``Semiglobal'' {\PLI}} 

Consider this \textit{semiglobal Polyak-\L{}ojasiewicz Inequality}:
\[
\redbox{(\forall \mbox{ compact } C\subset\Dom) (\exists \,\lambda _C>0)\;\norm{\gradloss(k)}^2 \geq  \lambda _C (\loss(k) - \lossmin)\;\; \forall \, k \in C}
\quad \red{(\sgPLI)}
\]
which implies
\[
\losst(t) \leq  e^{-\lambda _Ct} \losst(0)
\]
for all initial conditions in a sublevel set $C$.
It may well be the case that
$\lambda _C\rightarrow 0$ as $C\rightarrow D$,
so there are no guarantees of \textit{uniform} exponential convergence.
Since the optimal numbers $\lambda_C$ are nonincreasing as we take larger
sets, this indeed happens whenever there is no $\gPLI$ estimate. 

\minp{0.78}{%
For example, the diagram shows a case where, on the larger set $C_2$,
the rate of convergence may be very slow compared to the smaller set $C_1$:
\[
\losst(t) \leq  e^{-\lambda _{\green{C_1}}t} \losst(0), \;\;
\losst(t) \leq  e^{-\lambda _{\red{C_2}}t} \losst(0), \;\;
0 \approx \lambda _{\red{C_2}} < \lambda _{\green{C_1}}\,.
\]
}\minp{0.22}{%
\pic{0.25}{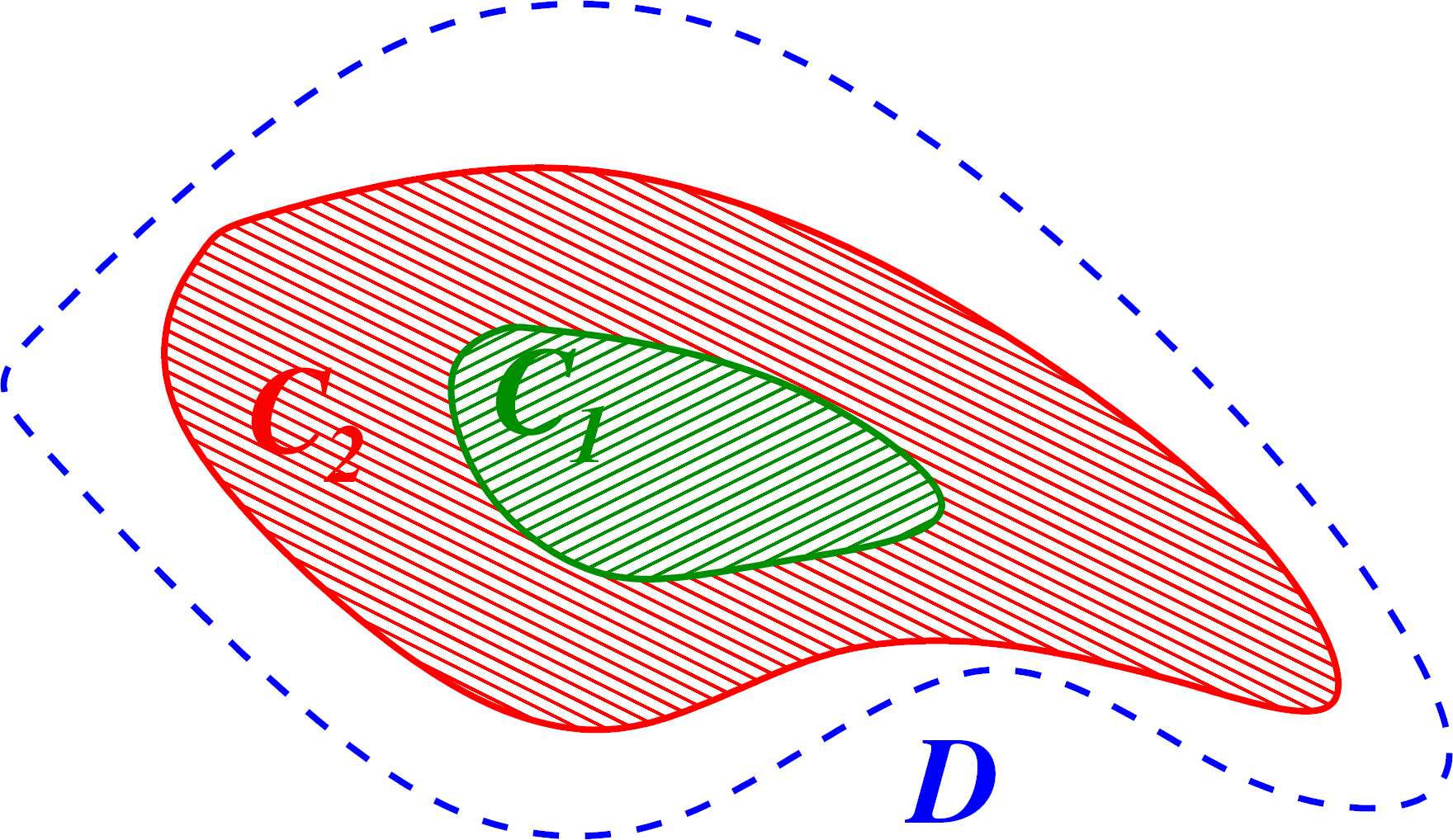}
}

\ms
This indeed happens for interesting problems, such as
the \textit{continuous time} LQR policy optimization
flow (though not for the \textit{discrete time} LQR problem).
We are then led to ask if there are global estimates of the form
$\norm{\gradloss(k)} \geq  \compfun (\loss(k) - \lossmin)$
where now $\compfun$ is a \textit{nonlinear comparison function}.
Furthermore, we will ask
how do the properties of $\compfun $ impact \textit{robustness to adversarial perturbations},
and explain how some of these properties guarantee a form of
``linear-exponential'' convergence.

\textit{Example:}
Let us revisit the loss for the LQR integrator example,
$n\!=\!m\!=\!1$,
$\dot x = u$, $x(0)=1$, $\int_0^\infty  x(s)^2 + u(s)^2\,ds$, $u(s)=-kx(s)$:
\[
\loss(k) \;=\; \frac{1+k^2}{2k}
\,,\;\; k\in \Dom=(0,+\infty ) \;\;\; \mbox{[$k$ stabilizing]}\,.
\]
Here
\[
\abs{\gradloss(k)}^2 \;=\;
\frac{1}{4} \left( 1 - \frac{1}{k^2} \right)^2 \;=\;
\twoif{\infty } {k\rightarrow 0}
      {1/4}{k\rightarrow \infty }
\]           
but, on the other hand,
\[
\loss(k) - \lossmin \;=\; 
\loss(k) - 1 \;=\;
\twoif{\infty } {k\rightarrow 0}
      {\infty }{k\rightarrow \infty }
\] 
and thus,
since $\gradloss(k)$ is bounded but $\loss(k)\rightarrow \infty$
as $k\rightarrow \infty $,
\textit{there is no possible estimate} of the form
$\norm{\gradloss(k)} \geq  \compfun (\loss(k) - \lossmin)$
where $\compfun$ is a continuous unbounded function, and in particular $\loss$
doesn't have the property $\gPLI$ ($\compfun(r) = \sqrt{\lambda r}$).
In fact, as for large $k(0)$ we have $\dot \losst(t) \approx -1/4$ we
should expect to see an approximately \textit{linear} decrease for
large initial conditions $k(0)$:
$\losst(t)\approx \losst(0)-t/4$, 
even though exponential convergence holds on each compact (\sgPLI).
%
%
%
Plotting an example,
we see an almost-linear convergence $\loss(k(t)) - \lossmin \rightarrow 0$,
followed by a ``soft switch'' to exponential decay:

%
%
%
\picc{0.17}{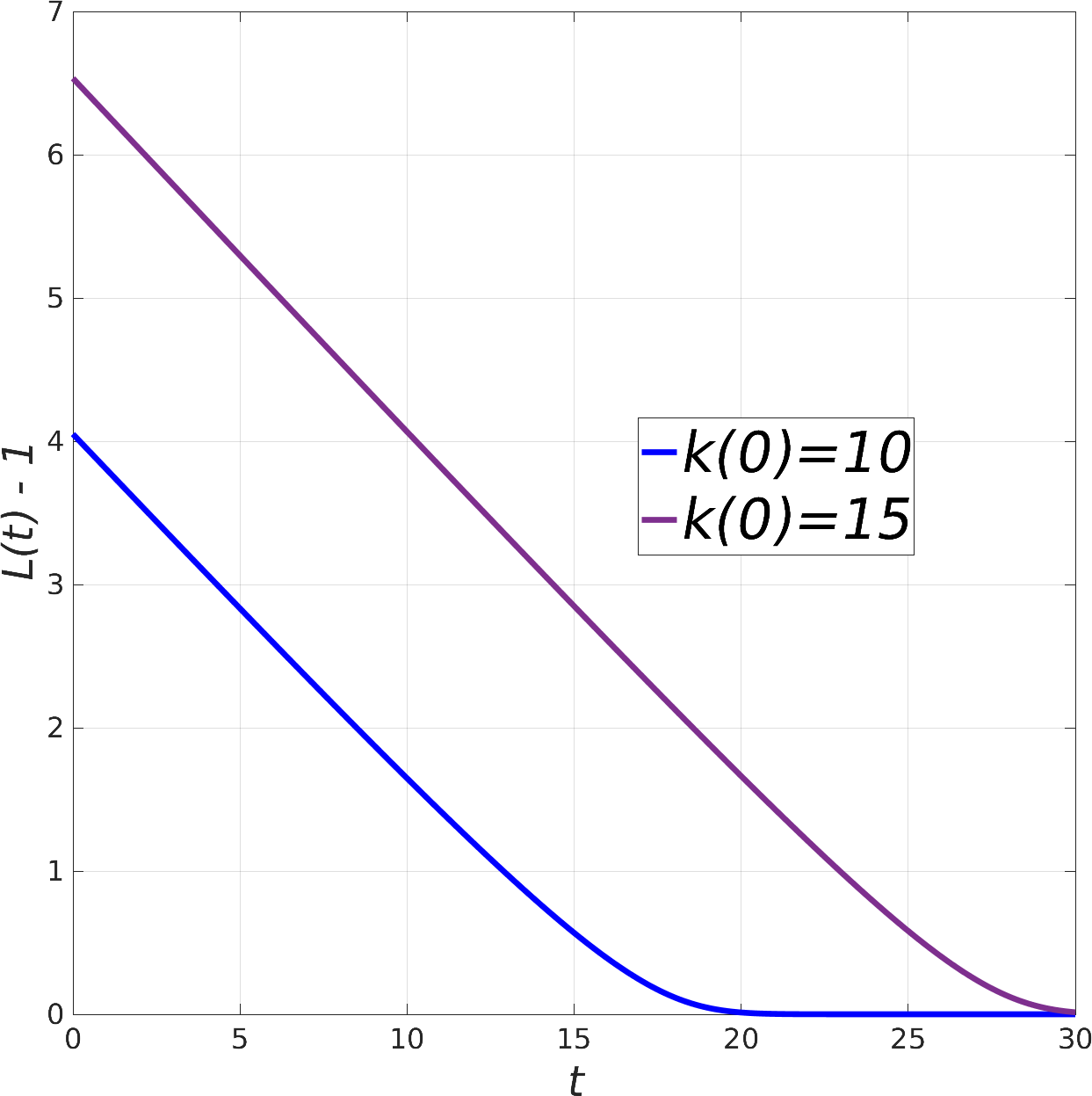}


%
%
%
%
%
%


\newsl{Weaker (but global) gradient dominance conditions}

Let us study several generalized forms of gradient dominance, all of the type:
\[
\redbox{\exists \,\lambda >0 \;\; \mbox{s.t.} \;\;\norm{\gradloss(k)}
  \geq
\alpha(\loss(k)-\lossmin) \;\; \forall \, k \in D}
\]
($\gPLI$ corresponds to $\alpha(r)= \sqrt{\lambda r}$).
These weaker versions will also guarantee \textit{(not necessarily exponential}) convergence,
but are more natural for the subsequent robustness (ISS-like) discussion.

We remind the reader that a continuous function $\alpha:\R_{\geq0}
\rightarrow \R_{\geq0}$ with $\alpha(0)=0$ is called positive definite,
class $\kk$, or class $\ki$ if these increasingly strong conditions hold:

\minp{0.35}{%

\brown{$\pd\,$}: $\,\alpha(r)>0$ for $r>0$,

\ms
\mbox{\green{$\kk\,$}: $\;\;\,$  if $\pd$ and also nondecreasing,}

\ms
\blue{$\ki\,$}:  $\,$ if $\kk$ and also unbounded.
}%
\minp{0.65}{\picc{0.5}{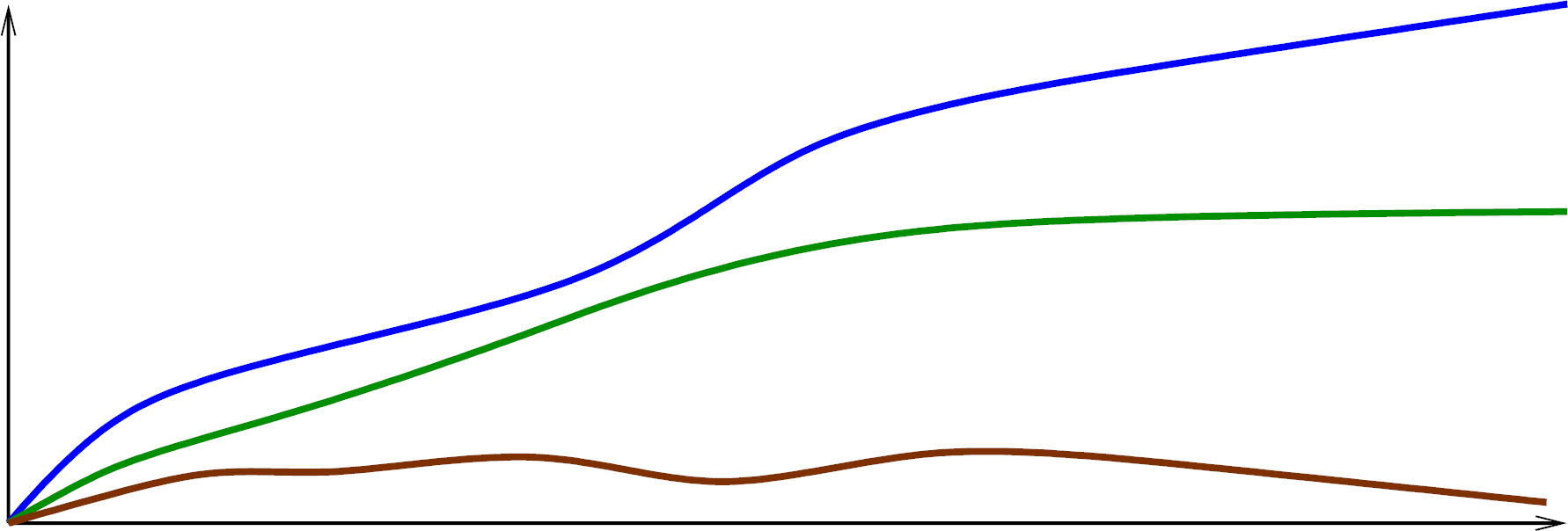}
}

\ms
Among $\kk$ functions, we wish to single out a new subclass,
which we call ``saturated'' \PLI (\satPLI), defined by the requirement that
there exist constants $a,b>0$ such that:
\[
{\alpha (r) \;=\; \sqrt{\frac{ar}{b+r}} \;\;\forall\, r\geq 0}\,.
\]
Observe that $\alpha(r) \approx \sqrt{\lambda r}$ ($\lambda = a/b$)
when $r = \loss(k)-\lossmin \approx 0$, which provides a \textit{local
  exponential} convergence, and $\alpha(r) \approx c = \sqrt{a}$ for
large $r$, which implies \textit{global linear} convergence.

Among $\pd$ functions are those of the form
\[
\sqrt{\frac{ar}{(b+r){\mathbf{^2}}}}
\]
(note the square) studied in
\cite{fatkhullin_optimizing_2021}
and other
papers on optimization.
Among $\ki$ functions that are not $\gPLI$, one has the
``Kurdyka-\L{}ojasiewicz'' estimates with  $\alpha^2$ strictly convex
\cite{fatkhullin2022sharp}.

We may also define local versions of these estimates, valid only for
$r\approx0$, but we only mention here $\lPLI$, a local version of $\gPLI$.

In summary, we have these classes:
\beqn
\kpl &:& 
\Big[\exists \, c>0 \Big]\;
\qquad\qquad\;\;\;
\alpha (r)\geq \sqrt{c\, r} \;\;\forall\, r\geq 0
\\
\phantom{a}&&
\\
\sgPLI &:& 
\Big[\forall\, \rho>0 \,,\, \exists \, c_\rho>0\Big] \;\;\;
\alpha (r)\geq \sqrt{c_{\rho } r} \;\; \forall\, r\in [0,\rho ]
\\
\phantom{a}&&
\\
\lPLI &:& 
\Big[\exists\, \rho>0 \,,\, \exists \, c_\rho>0\Big] \;\;\;
\alpha (r)\geq \sqrt{c_{\rho } r} \;\; \forall\, r\in [0,\rho ]
\\
\phantom{a}&&
\\
\phantom{aa} \ks &:& 
\Big[\exists \, a,b>0\Big] \;
\qquad\quad\;\;\;\,
\alpha (r)\geq \sqrt{\frac{ar}{b+r}} \;\;\forall\, r\geq 0
\eeqn
%
%
and the following implications:
\red{%
\[
\mymatrix{
\mbox{\gPLI} & \Rightarrow & \ks          & \Rightarrow & \mbox{\sgPLI}
& \Rightarrow & \mbox{\lPLI}
\\
\Downarrow  & \phantom{a} & \Downarrow    & \phantom{a} & \Downarrow
& \phantom{a} & \phantom{a} 
 \\
\ki         & \Rightarrow & \kk           & \Rightarrow & \pd\,.
& \phantom{a} & \phantom{a} 
}
\]%
}

These are important concepts, and next we will illustrate that fact by
analyzing the general LQR policy optimization.

\newsl{Results for the general LQR problem}

Consider the classical linear quadratic regulator problem for $\dot x=
Ax + B u$, for which the loss function is:
\[
{\loss(x_0,u(\cdot )) \,:=\; \int_0^\infty  x^T (t) \,Q\, x(t) \, + \, u^T(t)\, R\, u (t) \,dt}
\]
and which leads to the feedback solution
$u(t)=-\kfeedback_{\mbox{\small opt}}x(t)$,
with $\kfeedback_{\mbox{\small opt}} = R^{-1}B^T\pi$, where $\pi>0$
solves an associated algebraic Riccati equation.
To avoid dependence on initial states, one might assume a
(e.g.\ Gaussian) distribution over initial states $x_0$ and look at
the expected value of the cost.
In general, solving this problem requires the full knowledge of the
system and cost matrices $(A,B,Q,R)$.

\minp{0.80}{%
However, as emphasized by Levine and Athans as far back as 1970
\cite{1970levine_athans},
the problem can also be seen as a one of direct minimization of
a loss $\loss(k)$ in terms of the optimal $k$.
Now the gradient is must be estimated by an ``oracle'' which might
simulate a digital twin of a physical system or actually conduct experiments to
determine the loss of a particular policy.}%
\minp{0.20}{$\quad$ \pic{0.4}{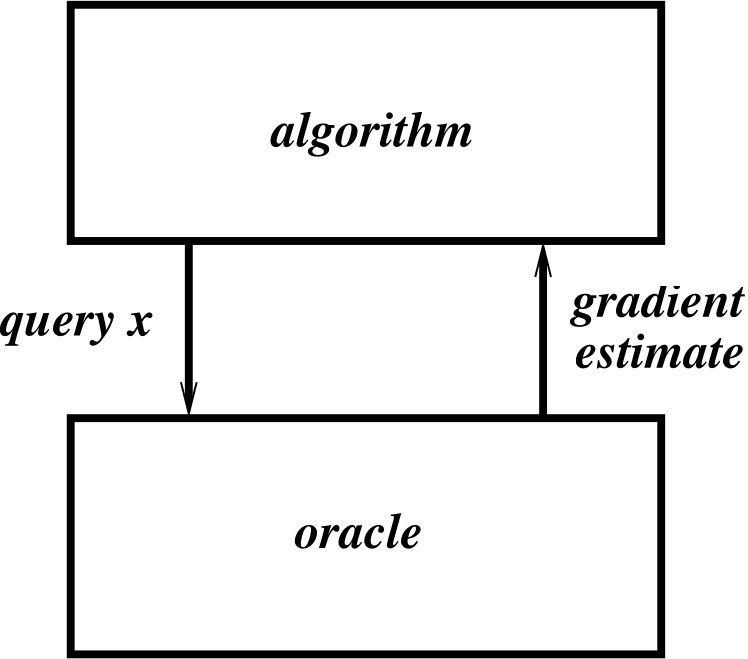}}%

As a search for $k$, the problem becomes:
\[
\min_{\kfeedback\in\Dom} \loss(\kfeedback), \; \mbox{where} \; \loss(\kfeedback) \, :=\; \Ex_{x_0}[\loss(x_0,\kfeedback \,\! x(\cdot ))]
\]
where $D$ is the open subset of $\R^{m\times n}$ of stabilizing
feedback matrices:
\[
\Dom = \Dom_{(A,B)}:=\{\kfeedback\stt A-B\kfeedback \;\mbox{Hurwitz}\}.
\]

\newsl{Why is the problem challenging?}

In general the problem
\[
\dot x = (A - B\kfeedback x), \;\;
\loss(\kfeedback):=\Ex_{x_0}\left[\int_0^\infty  x^T \left(Q + \kfeedback^T R \kfeedback\right) x \, dt\right]
\]
is not convex, and it leads to a complicated nonlinear gradient flow in matrix space:
\[
\dot k = -2(R\kfeedback\timearg - B^T P\timearg)Y\timearg
\]
where (if the covariance of initial states is the identity)
\[
(A-B\kfeedback\timearg)^TP\timearg + P\timearg (A-B\kfeedback\timearg) + Q + \kfeedback\timearg^TR\kfeedback\timearg=0
\]
\[
(A-B\kfeedback\timearg)Y\timearg + Y\timearg (A-B\kfeedback\timearg)^T + I_n=0 \,.
\]

\minp{0.55}{%
E.g.\ take $n$$=$$m$$=$$2$, 
$A$$ =$$ 0$, $B$$=$$Q$$=$$R$$=$$I$,
for which
$\kmin$$=$$I$.
The figure shows the intersection between $D$ and
the affine slice in $\R^{2\times2}$ consisting of matrices with the
following structure:
\[
\kfeedback =
\begin{bmatrix}
    1 &k_1 \\
   k_2 &1
\end{bmatrix}.
\]
(color map shows the level sets of the loss function).
This intersection, and hence $D$ itself, is clearly non-convex.
}%
\minp{0.45}{\picc{0.5}{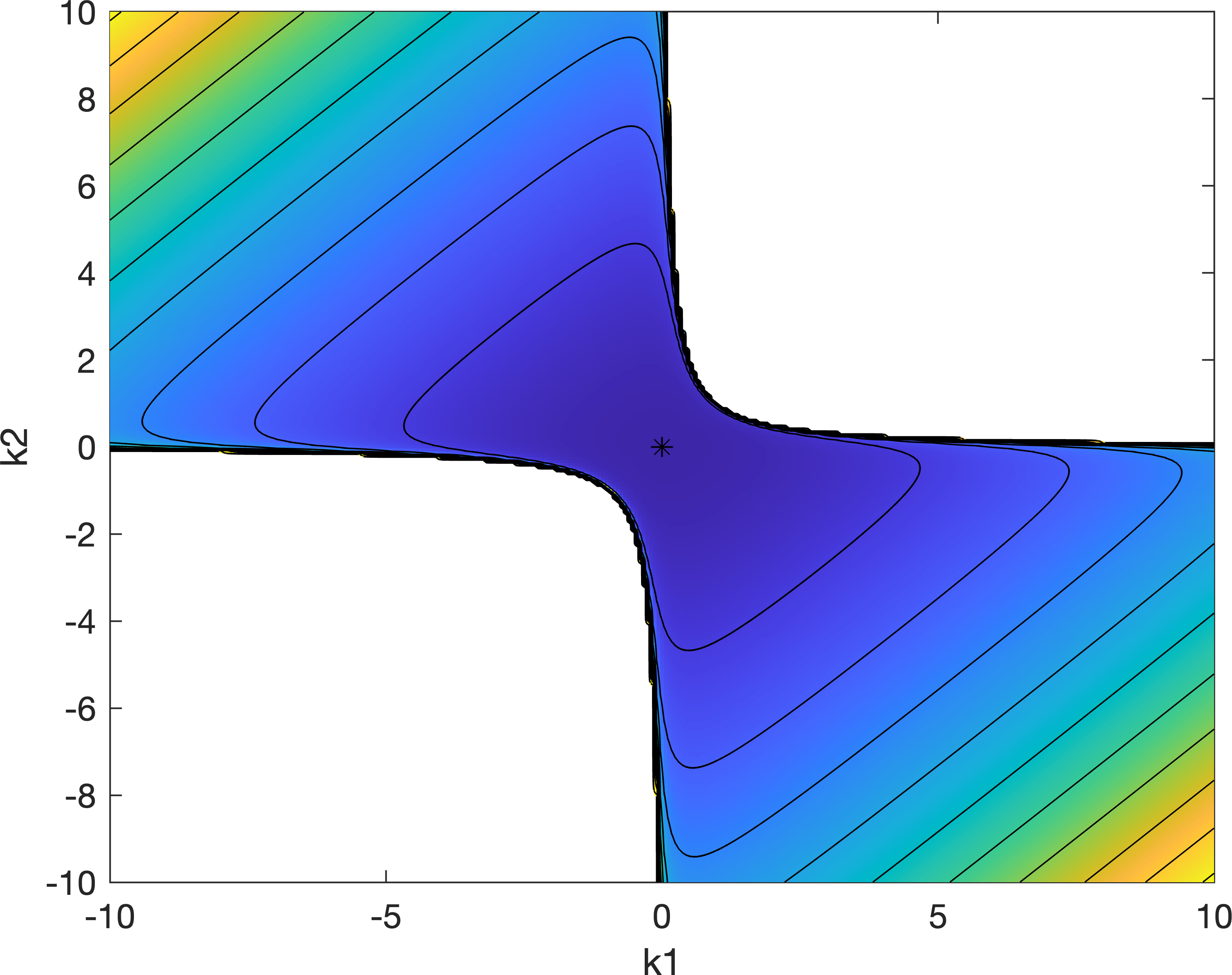}}

%
%

\newsl{A result for the general (continuous-time) LQR problem}

It is known that the loss of the LQR problem for any
\textit{discrete-time (DT)} system and any cost matrices:
\[
\loss(\kfeedback)\,:=\;
\Ex_{x_0}\left[
\sum_{t=0}^\infty x(t)^T\left(Q + \kfeedback^T R \kfeedback\right)x(t)\right],\;
\mbox{ where } x(t+1) = (A - B\kfeedback)x(t)\,, x(0)=x_0
\]
is of class $\gPLI$
\cite{fazel2018global}.
This guarantees global exponential convergence of the loss $\loss$ to its
minimal value, along solutions of the gradient system associated to a DT LQR problem.
In contrast, we already know from the one-dimensional integrator example
shown previously that the loss function for the continuous-time (CT)
LQR problem cannot generally be of class $\gPLI$.
It was only known for such problems that $\loss$ is of class $\sgPLI$ (thus also$\pd$)
\cite{bu2020policy,mohammadi_convergence_2022}.
This means that rates of exponential convergence may be vanishingly small
for initial conditions far from the minimizing feedback, and
convergence can at best be expected to display a linear-exponential
character as plotted earlier.
A recent result showed, however, that a $\satPLI$ condition does.
To be precise, consider the loss function
\[
\loss(\kfeedback)\,:=\;
\Ex_{x_0}\left[
\int_0^\infty  x(t)^T \left(Q + \kfeedback^T R \kfeedback\right) x(t) \, dt
\right],\;
\mbox{ where } \dot x(t) = (A - B\kfeedback)x(t)\, x(0)=x_0\,.
\]
Then we have:

{\bf Theorem.}
\cite{cui_jiang_sontag_2023_lqr}

\textit{$\loss$ admits a class $\satPLI$ (hence also $\kk$ and $\kplloc$) gradient dominance estimate.}

\newsl{Connection to $\gPLI$ in discrete-time LQR}

As we pointed out, \textit{global} {\PLI} holds for the
he loss $\loss$ associated to any \textit{discrete-time} LQR
problem, but not for the loss associated to continuous-time LQR problems.
Why the difference? To gain some intuition, let us use Euler's method to
discretize the example $\dot x=u$, $u=-kx$, $x(0)=1$, with step size $h>0$:
\[
x(t+h)\approx (1-hk)x(t) \;\;\Rightarrow \;\;
x(ih) \approx (1-hk)^i
\]
which leads to the following loss function:
\[
\loss_h
\;\approx\;
\sum_{i=0}^\infty  h \left[ [x(ih)]^2 + [-kx(ih)]^2 \right]
\;=\;
\frac{1+k^2}{k(2-kh)}\,.
\]
It can be shown that this is the correct loss, in the sense that it
produces the same optimal feedback as the CT problem as
$h\rightarrow0$.
(Note the ``$h$'' factor which arises from the approximation of the
integral of the constant $x(ih)$ over each interval $[ik, (i+1)h]$.)
The domain of $\loss_h$ is $D_h=(0,2\!/\!h)$ (so that the closed loop
system is stable, $\abs{1-hk}<1$).

For any \textit{fixed} $h$, we have $\loss_h(k)\rightarrow \infty $ as
$k\rightarrow 0,2\!/\!h$, and also $\abs{\loss_h'(k)}\rightarrow\infty $,
so it is immediate that there is some function $\compfun _h\in \ki$ such that
\[
\norm{\gradloss_h(k)} \geq  \compfun _h(\loss_h(k) - \min_k \loss_h(k)).
\]
To gain more insight, one can show that, in fact, one can get a
$\gPLI$ estimate in which $\compfun _h(r) = \sqrt{\lambda_h r}$ with 
$\lim_{h\rightarrow 0}\lambda _h = 0$.
In other words, the estimate becomes useless as $h\rightarrow 0$,
consistently with there not being a $\gPLI$ estimate for the CT LQR
problem.

\section{Perturbed gradient systems and {\PLI} variants}

In practice, one may expect that the gradient $\gV$ might be
imprecisely computed by an ``oracle'' physical or numerical simulator.

\minp{0.5}{%
Disturbances or errors may arise from, for example:

\itt 
adversarial attacks,

\itt
errors in evaluation of ${\cal L}$ by oracle,

\itt
early stopping of a simulation,

\itt
stochastic computations (``reproducibility''),

\itt
inaccurate and very approximate digital twin, or

\itt
learning by sampling from limited data.
}%
\minp{0.32}{\picc{0.55}{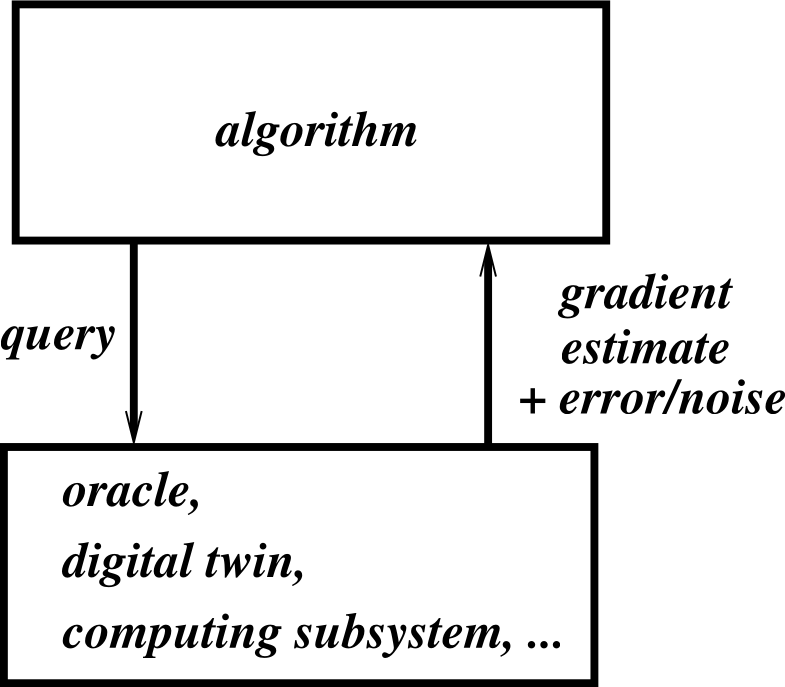}}%
\minp{0.20}{\
\phantom{a}
\bigskip

\pic{0.2}{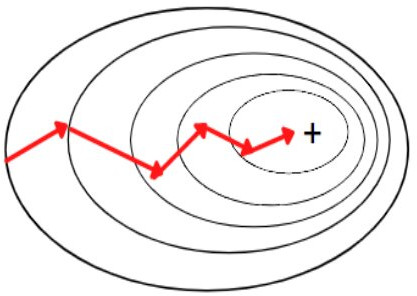}
}

\ms
We will use the simplest model of such ``error'' or ``disturbance''
inputs $u(t)$ as additive terms:
\[
\dot \argL(t) \;=\; f(\argL(t),u(t))
\;=\;
-\gV(\argL(t))^\top + \,B(\argL(t))\,u(t) 
\]
where $B:\X\rightarrow \R^{n\times m}$ is a bounded locally Lipschitz function.

Such perturbed gradient systems have been studied for a long time
(e.g.~\cite{polyak1987introduction,bertsekas1999nonlinear,bertsekas2000gradient}),
but we take here a different point of view, that of
\textit{input to state stability (ISS)}.
ISS is a property which encapsulates the
idea of a ``graceful degradation'' (transient and asymptotic) of
convergence if the disturbance or error input $u$ is in some sense
``small'' (transient, asymptotically, or in some average sense).
In this context, we would like that
$\dist(\argL(t),\target)\approx0$, or at least
$\V(\argL(t))\approx\lossmin$, for large times $t$,
and an estimate of the rate of convergence assuring that there is
not much ``overshoot'' in $\dist(\argL(t),\target)$ or $\V(\argL(t))$
for small $t$.

\newsl{Review: ISS-like properties on open sets}

%
%
%
%
%
%
%
%
%
\minp{0.75}{%
The ISS property is typically studied in all of $R^n$. However, when
applied to gradient flows we need to restrict the domain
(e.g.\ to stabilizing feedbacks), so we need to define properties 
relative to an open subset $\target \subset \X\subseteq \Rn$.
We can do this as follows.
A function
$\omega : \X \rightarrow  \R$ is a ``barrier metric'' or \textit{size
  function for} $(\X,\target )$ if 
it is
continuous, positive definite with respect to $\target $, and proper
(or ``coercive''), meaning that $\omega (k)\rightarrow \infty $ as
$k\rightarrow \dX$ or $\abs{k}\rightarrow \infty $).}%
\minp{0.25}{\null\phantom{aaa}\pic{0.42}{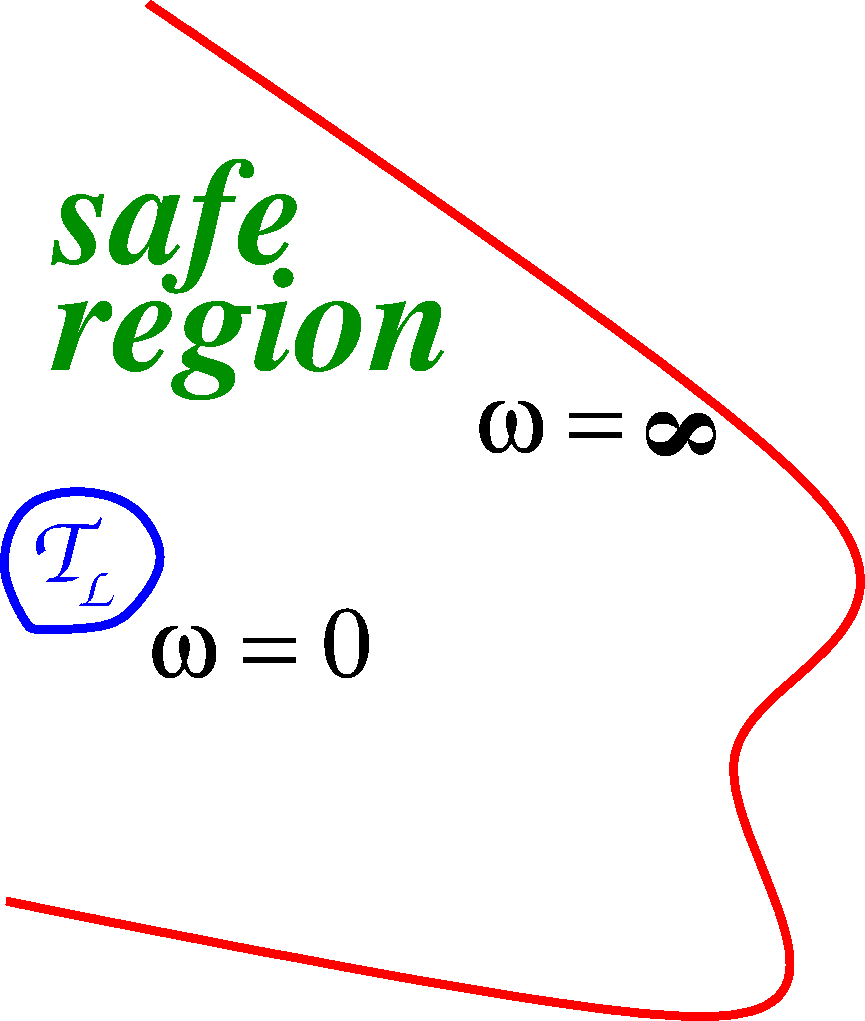}}


\newsl{Review: [integral, small-input] input to state stability ([i, si] ISS)}

%
%
%
%
%
%
%


\minp{0.6}{%
  The system $\dot \argL(t) \!=\! f(\argL(t),\red{u(t)})$
  is \textit{ISS} if
there exist $\beta\! \in\! \kl$ and $\ggt\!\in\! \ki$ such that, for all
initial conditions $\argL(0)$ and for all (Lebesgue measurable and
essentially bounded) inputs $u$,
\[
\omega(\argL(t))
\;\leq \;\max
\left\{
\beta (\omega(\argL(0)),t) \,,\,
\ggt\left(\normi{u}\right)
\right\}
\;\forall \; t\geq0\,.
\]
For small-input ISS, we only require that there is some $M>0$ such
that the estimate holds for $\normi{u}\leq M$.
For integral ISS, the input term has the form
$\int_0^t \ggt(\abs{u(s)}) \,ds$.
}%
\minp{0.3}{%
\setlength{\unitlength}{2800sp}%
\begin{picture}(8424,2424)(7000,-4000)
  \thicklines
\put(7201,-3961){\vector( 0, 1){2400}}
\put(7201,-3961){\vector( 1, 0){3600}}
\qbezier(7201,-2536)(8026,-1936)(8626,-2536)
\qbezier(8626,-2536)(9226,-3136)(10426,-3436)
\put(7930,-2275){\red{\vector( 0, 1){  0}}}
\put(7930,-2275){\red{\vector( 0,-1){1650}}}
\put(7296,-2086){\red{\small  overshoot $\leq \beta(\omega(\argL_0),0)$}}
\put(10801,-4300){$t$}
\put(7351,-1711){$\omega(\argL(t))$}
\put(10501,-3511){\red{\vector( 0, 1){  0}}} 
\put(10501,-3511){\red{\vector( 0,-1){375}}}
\put(9600,-2900){\red{asymptotically}}
\put(10000,-3200){\red{$\leq\ggt\left(\normi{u}\right)$}}
\end{picture}
}

\ms
Recall that $\beta\in\kl$ means that
$\beta(\cdot,t)\in\ki$ for every $t$, and
$\beta(\cdot,t)\rightarrow0$ as $t\rightarrow\infty$.
(One may always assume, without loss of generality, that $\beta(r,t)$
is of the form $\alpha_1( e^{-\lambda t} \alpha_2(r))$ for some $\alpha_i\in\ki$.


\newsl{Review:~[i]ISS is natural generalization of linear stability}

For linear systems $\dot x=Ax+Bu$ (with Hurwitz matrix $A$), typical
estimates of stability for operators:
\[
\left\{L^2,L^\infty \right\} \rightarrow  \left\{L^2,L^\infty \right\}\,:\; (x_0,u) \mapsto  x(\cdot )
\]
are:
\beqn
\abs{\xtou} &\leq &
c_1 \abs{\xo}e^{-\lambda t} \,+\,
c_2 \sup_{s\in [0,t]} \abs{u(s)}
\qquad(L^\infty \rightarrow L^\infty )
\\
\abs{\xtou}  &\leq &
c_1 \abs{\xo}e^{-\lambda t} \,+\,
c_2  \int_0^t \abs{u(s)}^2 \,ds
\quad\;(L^2\rightarrow L^\infty )
\\
\int_0^t \abs{\xsou}^2 \,ds\, &\leq &
c_1  \abs{\xo}
\quad
\quad
\,+\,
c_2  \int_0^t \abs{u(s)}^2 \,ds
\quad\;\;(L^2\rightarrow L^2)\eeqn
(the missing case $L^\infty  \rightarrow L^2$ is less interesting, being too restrictive).
For linear systems, all these notions turn out to be equivalent (with
different constants). 
For nonlinear systems, and under (nonlinear) changes of coordinates
$x(t)=T(z(t))$, one arrives to the notions of ISS, iISS, and (again)
ISS respectively.
In that sense, ISS and iISS are direct generalizations of standard
linear stability estimates.

\newsl{Review: dissipation inequalities}

A continuously differentiable function 
$\V:\X\rightarrow \R$ is a (dissipation form)
\textit{ISS-Lyapunov function} for $\dot \argL=f(\argL,u)$ with
respect to $(D,\target)$ if these two properties hold:
\bi
\item
($\exists\,\lossmin$)$\;$ $\V-\lossmin$ is a size function for
  $(D,\target)$, and
\ms
\item
($\exists$ $\alpha ,\gamma \in \ki$)$\;$
$
\dot  \V(\argL,u) \; \leq \; -\alpha (\V(\argL)-\lossmin)\,+\,\gamma (\abs{u})
\quad\forall\, (\argL,u)\in \X\times \R^m
$,
\ei
where $\dot \V(\argL,u) := \nabla \V(\argL)\cdot f(\argL,u)$, which
means that $d\V(\argL(t))/dt = \dot \V(\argL(t),u(t))$ along solutions of $\dot \argL=f(\argL,u)$.

An \textit{iISS-Lyapunov function} is one for which the second
property is satisfied with $\alpha \in \kk$, and a
\textit{siISS-Lyapunov function} is one for which the second property
is satisfied for inputs satisfying $\abs{u}\leq M$ for some $M>0$.

Such Lyapunov functions are key to studying ISS-like properties.
For compact target sets $\target$:

\textbf{Theorem.} There exists an \{ISS,iISS,siISS\} Lyapunov function
if and only of the system is \{ISS,iISS,siISS\}.

Compactness is only needed in order to prove necessity;
the implication
\ms

\centerline{exists \{ISS,iISS,siISS\}-Lyapunov function $\Rightarrow$  system is \{ISS,iISS,siISS\}}

is always true. In fact, the converse is also true in the
noncompact case, provided that the definition is
slightly modified to what is called a ``conditional'' form, in which
decay of $\loss$ is only required for large states.
These properties are well-studied and by now classical.
See for instance~\cite{04cime}.

When applied using $\loss$ as a Lyapunov function, one obtains the
following very elegant and concise characterizations of the behavior
of perturbed gradient systems:

%

\textbf{Theorem.}
Suppose that $\V$ is a size function with respect to
$(D,\target)$. Then the following properties hold for the system
$\dot k(t) = -\gV(\argL(t))^\top + B(\argL(t))\,u(t)$:
  
\bi
\item
$\norm{\gV(\argL)} \; \geq  \; \alpha (\V(\argL)-\lossmin)$, for some $\alpha \in \ki$
$\quad\,\implies\quad$
the system is ISS
\item
$\norm{\gV(\argL)} \; \geq  \; \alpha (\V(\argL)-\lossmin)$, for some $\alpha \in \kk$
$\quad\quad\implies\quad$
the system is siISS
\item
$\norm{\gV(\argL)} \; \geq  \; \alpha (\V(\argL)-\lossmin)$, for some $\alpha \in \pd$
$\quad\,\implies\quad$
the system is iISS.
\ei
We summarize with the following diagram:

\red{\[
\mymatrix{
\mbox{\gPLI} & \Rightarrow & \ks          & \Rightarrow & \mbox{\sgPLI}
& \Rightarrow & \mbox{\lPLI}
\\
\Downarrow  & \phantom{a} & \Downarrow    & \phantom{a} & \Downarrow
& \phantom{a} & \phantom{a} 
 \\
\ki         & \Rightarrow & \kk           & \Rightarrow & \pd
& \phantom{a} & \phantom{a} 
 \\
\Downarrow& \phantom{a}  & \Downarrow & \phantom{a} & \Downarrow
& \phantom{a} & \phantom{a} 
\\
\mbox{ISS}  & \phantom{a}  & \mbox{siISS} & \phantom{a} & \mbox{iISS}
& \phantom{a} & \phantom{a} 
}
\]
}

Note the following consequence: if $\loss$ satisfies a $\satPLI$ estimate, then
the perturbed gradient flow is both iISS and siISS, a property sometimes called
``strong'' iISS''.

\newsl{Application to LQR problem}

Our work in this area was in fact motivated by trying to prove the
following result, which is an immediate application of the above:

\textbf{Theorem.}
\cite{cui_jiang_sontag_2023_lqr}

The perturbed gradient flow associated to any (continuous-time) CT LQR
problem is strongly iISS.\ms

This greatly generalizes previous a result
\cite{sontag2022iss_gradient_flows}
which only proved iISS, as that result was
based on the previously known $\sgPLI$ estimate.

We note that \cite{cui_jiang_sontag_2023_lqr}
also derived similar properties for perturbed natural gradient flows
\[
\dot \kfeedback(t) = - 2\eta (R\kfeedback(t) - B^TP(t)) + u(t)
\]
where $2 (R\kfeedback(t) - B^TP(t))$ is the gradient over the
Riemannian manifold
($D$, $\langle \cdot,\cdot\rangle_{Y}$) 
and $Y$ is the solution of a closed-loop Lyapunov equation. 
Also given there is an siISS result for a Gauss-Newton flow
\[
\dot \kfeedback(t) = -\eta(\kfeedback(t) - R^{-1}B^TP(t)) + u(t)\,.
\]
Finally, we note that the results can be adapted to show ISS-like
properties for gradient descent (iterative, discrete steps)
\cite{sontag2022iss_gradient_flows,%
cui_jiang_sontag_2023_lqr}

%
%
%
%
%
%
%
%
%
%
%
%

\newsl{Remark: other works on ISS-like gradient flows}

There have been several papers that studied perturbed gradient flows
from an ISS(-like) viewpoint, though apparently the key connections with
different variants of {\PLI} estimates had not been remarked upon.
A partial list is as follows:

\itb
   Saddle dynamics
\cite{ieee_tac_2018_cherukuri_et_al_convexity_saddle_point_dynamics}
  (ISS gradient with respect to additive errors, $V$ has a convexity property, $\X=\Rn$)

\itb
  Fixed-time convergence in extremum seeking
  \cite{2021_poveda_krstic_fixedtime_iss_extremum_seeking}
(gradient flow, ``D-ISS'' property with respect to a time-varying uncertainty, $\X=\Rn$)

\itb
  Switched {LTI} systems
  \cite{2020arxiv_bianchin_poveda_dallanese_gradient_iss_switched_systems}
   (ISS gradient flow with respect to unknown disturbances acting on plant, $\X=\Rn$)

\itb
  Extremum seeking
\cite{2021_arxiv_iss_gradient_suttner_dashkovskiy}
(ISS gradient flow for kinematic unicycle, $\X$ closed submanifold of $\Rn$)

\itb
  Bilevel optimization
\cite{kolmanovsky2022inputtostate}
(ISS prox-gradient flow; errors arise from ``inner loop'' incomplete optimization)

\itb
  LQ 
\cite{Pang2022}
(Kleinman’s policy iteration, ``small-input'' ISS).

\section{(Linear) neural net controllers}

We next summarize some recent results obtained with Arthur de Olivera and others
(\cite{24submitted_arthur_nn_gradient,25moh_arthur_eduardo_in_preparation}) regarding implementations
of optimal LQR controllers using feedforward neural networks.

\minp{0.64}{%
A feedforward neural network is a function $u=k(x)$ that can be expressed as a
composition of functions:
\[
y = (D_N \circ \KK_{N}\circ D_{N-1}\circ \KK_{N-1}\circ \ldots \circ D_1 \circ \KK_{1}\circ D_{0})(u)
\]
where the maps $\KK_i$ are linear and the $D_i$ are coordinatewise
applications of a fixed scalar ``activation'' function (typically
$\tanh$ or a rectified linear unit ReLU).

For our study of the LQR problem
we take the activation function to be
the identify (the $D_i$'s are identity maps):
\[
u = \KK_N \KK_{N-1} \dots \KK_2\KK_1 x = \KK x\,.
\]
}%
\minp{0.36}{%
\picc{0.18}{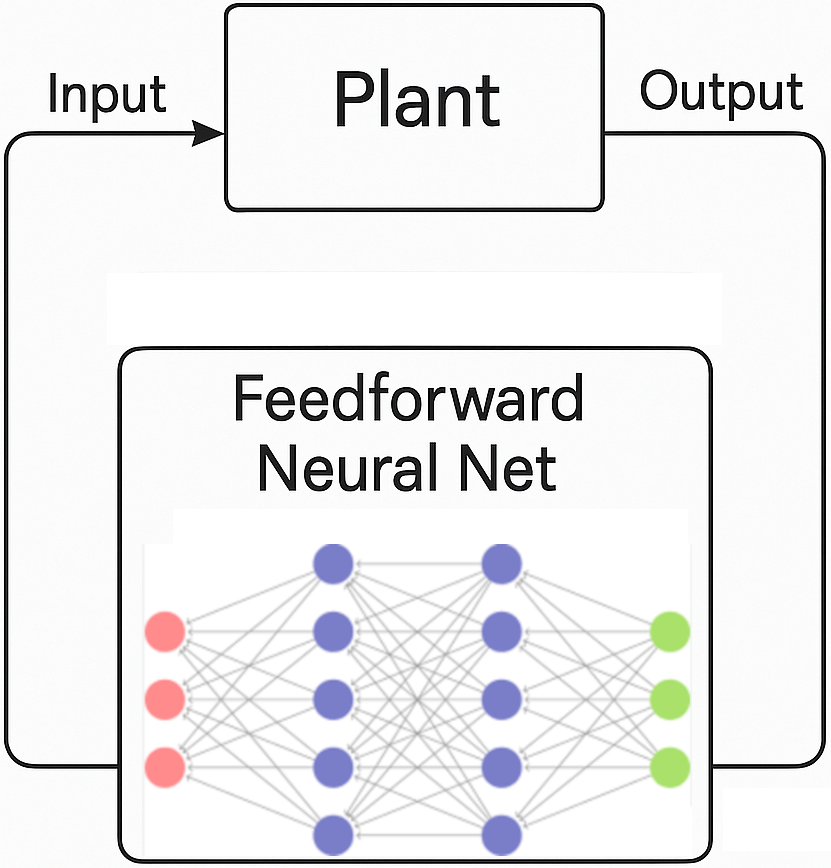}}

\ms
We call the search for feedback of the form
$\KK_N \KK_{N-1}\dots\KK_2\KK_1$ an ``overparametrized''
problem, because one now looks for solutions having a very
large number of parameters, For example, if inputs $u$ and
states $x$ are one-dimensional, then $u=\KK x$ has just one
parameter to be found, but if $N=2$ and the matrices $\KK_2$
and $\KK_1$ have size $1\times\kappa$ and $\kappa\times1$
respectively, then there are $2\kappa$ parameters to be found.

Why studying this problem? First of all, because it is of interest to
ask what happens if a neural network approach is used ``blindly''
without knowing that there is a linear feedback solution. Of course,
in typical applications of neural network control, activation
functions are nonlinear, but the linear case is natural for this
applications, is easier to analyze, and it provides very useful
intuition. Second, and perhaps surprisingly, it will turn out that
gradient systems searching for the optimal feedback converge
``faster'' (in a certain sense) in the overparametrized formulation that
the simple one-matrix optimization.

\newsl{Digression on neural nets}

The topic of neural networks is of course a very old one.
Artificial neural networks were proposed in 1943 by McCulloch
and Pitts (``A logical calculus of the ideas immanent in
nervous activity'').
By the late 1950s Rosenblatt had shown how ``perceptrons'' could ``learn''
basic functions, and actual physical machines were built by the US
Navy at the time. As quoted in the NY Times (8 July 1958), this was
supposed  to lead to a computer ``that [the Navy] expects will be
able to walk, talk, see, write, reproduce itself and be conscious of
its existence.''

Personally, I was interested on the topic way back, as a mathematics
undergraduate in Argentina, and published in 1972 a book on the
subject which surveyed work being done at the time (including by an AI research
group in Buenos Aires) on pattern recognition, stochastic gradient
descent for reinforcement learning from rewards and penalties, and the
idea that training data comes with probability distributions.

\minp{0.5}{\picc{0.4}{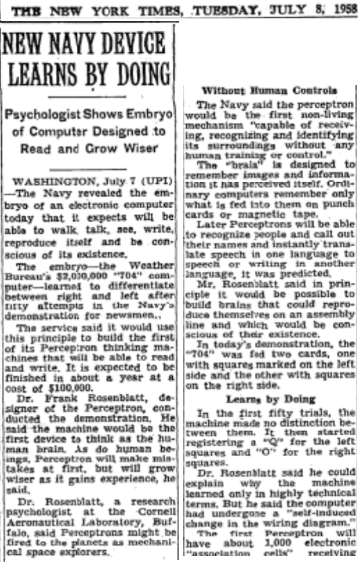}}
\minp{0.5}{\picc{0.4}{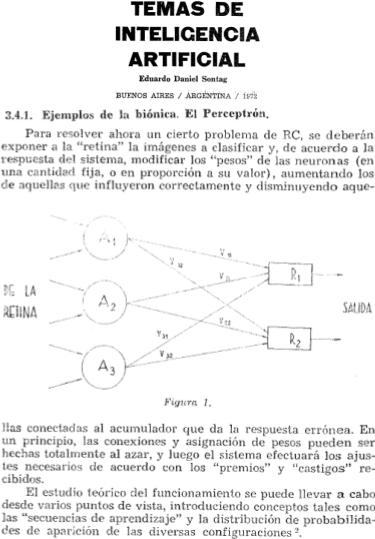}}

Of course, current developments in hardware (GPU's, in particular),
availability of huge amounts of data for training/learning, and novel
architectures (such as attention mechanisms and transformers) have
created a true revolution in capabilities.

We formulate the following optimization problem:
\[
\min_{\K\in\Kdom}
\loss(\K) =
  \mathbb{E}_{x_0\sim\mathcal{X}_0}
   \left[\int_0^\infty x(t)^\top Q x(t) + u(t)^\top R u(t)\, dt\right]
\]
for a linear system $\dot x = Ax+Bu$ and feedback in the following product form:
\[
u \,=\, \K\, x \, =\, {\KK}_N\dots {\KK}_2{\KK}_1x
\]
where ${\KK}_1$$\in$$\R^{\kappa_1\times n}, {\KK}_2$$\in$$\R^{\kappa_2\times \kappa_1}, \cdots, {\KK}_N$$\in$$\R^{m\times \kappa_{N-1}}$,
We call the tuple $\K = ({\KK}_1,{\KK}_2,\dots, {\KK}_n)$
a ``Linear Feedforward Neural Network (LFFNN) with $N\!-\!1$ hidden layers.''
The set $\Kdom$ consists of those tuples $\K$ for which 
the product ${\KK}_N{\KK}_{N-1}\dots {\KK}_1$ stabilizes,
and study the study gradient flow:
\[
\dot {\KK}_i = - \nabla_{{\KK}_i}\loss(\K)
\]
initialized at $\K\in\Kdom$

\newsl{Well-posedness theorems}

We have the following existence and convergence results:

\itb solutions exist for $t\geq0$,

\itb are precompact,

\itb remain in the set $\Kdom$ of stabilizing controllers, and

\itb converge to critical points of the gradient flow dynamics.

Furthermore, in the single hidden layer case ($N=2$), we
have the following additional properties:

\itb every equilibrium $({\KK}_1,{\KK}_2)$ is either a global minimum
or a strict saddle,

\itb the product $\widehat{\K} = {\KK}_2{\KK}_1$ at critical points
corresponds to low-rank approximations of the optimal $\K^*$, and

\itb except for a zero-measure set of initializations, solutions
are such that the product $k_2k_1$ converges to
the optimal feedback $\K^*$.

The low-rank approximation property means that there exists a singular value
decomposition of $\K^*$:
\[
\K^* \;=\; 
U
      \begin{bmatrix}
            \Sigma_1^* & 0 \\ 0 & \Sigma_2^*
      \end{bmatrix}
V^\top
\]
such that (with nonzero $\Sigma_1^*$, and after reordering if necessary):
\[
\widehat{\K} \;=\; {\KK}_2{\KK}_1 \;=\; 
U       \begin{bmatrix}
           \Sigma_1^* & 0 \\ 0 & 0
        \end{bmatrix}
.
V^\top
\]
In particular, the number of critical values of $\loss$ is finite
(evaluate at the possible $\widehat{\K}$'s), thus partitioning critical points 
into finitely many algebraic varieties of dimension $(\kappa-r)(n+m)+r^2$
(empty if $\kappa<r=\mbox{rank}\,\widehat{\K}$; embedded submanifold if $\kappa=r$).

%
%
%

\newsl{Imbalance}

Analogously to similar notions for linear neural networks in
regression problems, there are constant matrices $ \mathcal{C}_i$ such
that, along any solution of the gradient system and all $t\geq0$, and $i\in\{1,\ldots,N-1\}$:
\[
{\KK}_i(t)\,{\KK}_i(t)^\top - {\KK}_{i+1}(t)^\top {\KK}_{i+1}(t)
\;\equiv\; \mathcal{C}_i\,.
\]
Let us introduce for each $i$:
\[
c_i \,:=\;\sqrt{\abs{\sum \lambda_j(\calC_i)^2 -2 \sum_{j < k} \lambda_j(\calC_i) \lambda_k(\calC_i)}}
\]
which is an eigenvalue concentration measure of imbalance.

The main results say, in very informal terms, that
the larger the imbalance, the faster the convergence to the
optimum, and more robustness (in an ISS sense).
We have verified these properties numerically and developed
theoretical results for the special case $n\!=\!m\!=\!1$ (and $N\!=\!2$)
\cite{24submitted_arthur_nn_gradient,25moh_arthur_eduardo_in_preparation}.
Here we just show a simple set of examples, to develop the intuition.

\newsl{Theory for $n\!=\!m\!=\!1$, $N\!=\!2$ (but arbitrary $\kappa_1$)}

Without loss of generality, we consider systems  $\dot x = ax + u$,
i.e. set $b=1$.
The gradient dynamics takes the following explicit form:
\beqn
    \dot k_1 &=& f(k_2k_1)\,k_2^\top\\
    \dot k_2 &=& f(k_2k_1)\,k_1^\top
\eeqn
where
\[
f(k)\;=\; \frac{rk^2-2ark-q}{2(a-k)^2}\,.
\]
This is simply a linear saddle, but with an inversion of the flow on
the set of pairs $(k_1,k_2)$ where $f(k_1k_2)<0$, and new equilibria
at points $f(k_2k_1)=0$. 

\minp{0.4}{\pic{0.48}{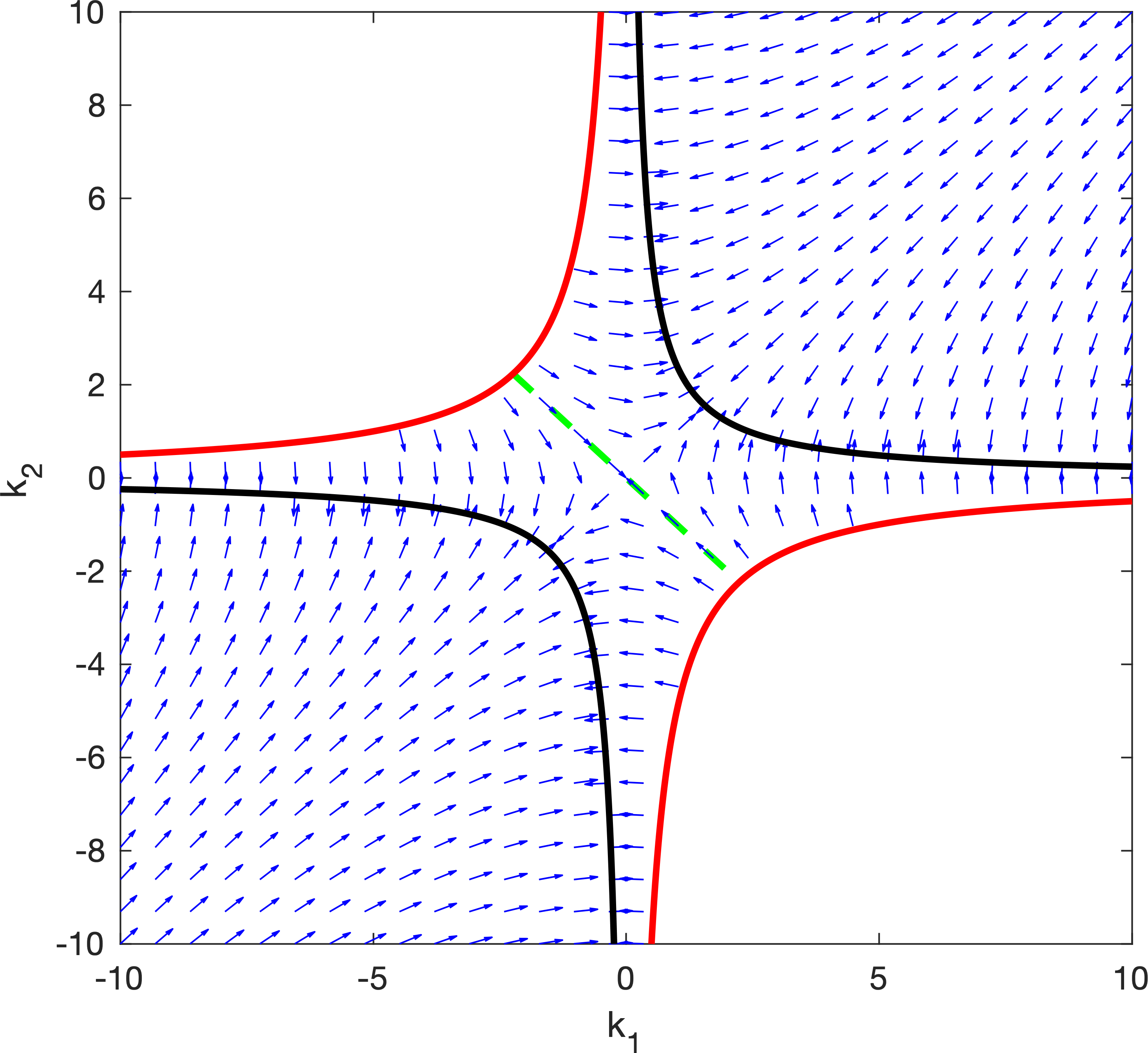}}%
\minp{0.6}{%
For example, this is the phase plane for $\kappa_1\!=\!1$ (with $a<0$).

\blue{Blue arrows} show the vector field.

Black hyperbolas show the equilibria, which is the set of pairs
$(k_1,k_2)$ such that $f(k_2k_1)=0$
(i.e. $k_2k_1 = a+\sqrt{a^2+q/r}$).

\red{Red hyperbolas} show the boundary of the stabilizing set
(the vector field is only defined there).

\green{The green dashed line} is the stable manifold of the saddle,
$k_1=-k_2^\top$ (restricted to the domain).
}

One can show that $\K(t) = (k_1(t),k_2(t))$ converges to target set
(optimal feedback) iff $k_1(0)\!+\!k_2(0)^\top\not=0$,
i.e., except from stable manifold of the unique saddle at $(0,0)$.

\textbf{Imbalance Speed-up Theorem:}
Suppose that two solutions have:
%
%
%

\itb
  the same initial cost: $\loss(\tilde{\K}(0))=\loss(\bar{\K}(0))$, but

\itb  
  the respective imbalances satisfy: $\tilde c>\bar c>0$,

Then, $\loss(\tilde{\K}(t))<\loss(\bar{\K}(t))$ for all $t>0$.
In other words, the cost converges faster when initialized with a larger
level of imbalance.

\newsl{Rates: recovering \gPLI in the overparametrized case}

In the (continuous-time) LQR problem, we only had \sgPLI and \satPLI
estimates, but there is no \gPLI estimate. However, it turns out that the
overparametrized problem allows one to recover \gPLI estimates on
uniformly imbalanced forward-invariant sets for the gradient system
$\dot {\KK}_i = -\nabla_{{\KK}_i}\loss(\K)$, leading to global
exponential convergence and ISS-like properties.
Obviously, the forward invariant set cannot include spurious critical
points, because even a local {\PLI} estimate will give that
$\nabla\loss(\K)\!=\!0$ implies $\loss(\K)\!=\!\lossmin$. Thus we need
to avoid such points.

General results for $n\!=\!m\!=\!1$ and $N\!=\!2$ are shown
in \cite{25moh_arthur_eduardo_in_preparation}, but let us sketch a simple
example here.
We let $q\!=\!r\!=\!1$, and $a\!=\!-1$.
One can show that for some constant $\rho>0$:
\[
\norm{\nabla \loss(\K(t))}^2 \;\geq\; \rho\, 
\min\left\{\norm{k_1}^2 +\norm{k_2}^2,1\right\}
\,\left(\loss(\K(t))-\lossmin\right)
\]
and then note that, on forward invariant sets made up of solutions with
imbalance 
$ c \geq \sqrt{\mu}$, it holds that 
$\norm{k_1}^2 + \norm{k_2}^2 \geq c^2$,
so we have a \gPLI estimate:
\[
\norm{\nabla \loss(\K(t))}^2 \;\geq\; \tilde\rho \,\left(\loss(\K(t))-\lossmin\right)
\]
where $\tilde\rho:=\rho\, \min\left\{\mu,1\right\}$.
This means that there is a \gPLI estimate on regions where the imbalance is bounded below.

\minp{0.43}{\picc{0.35}{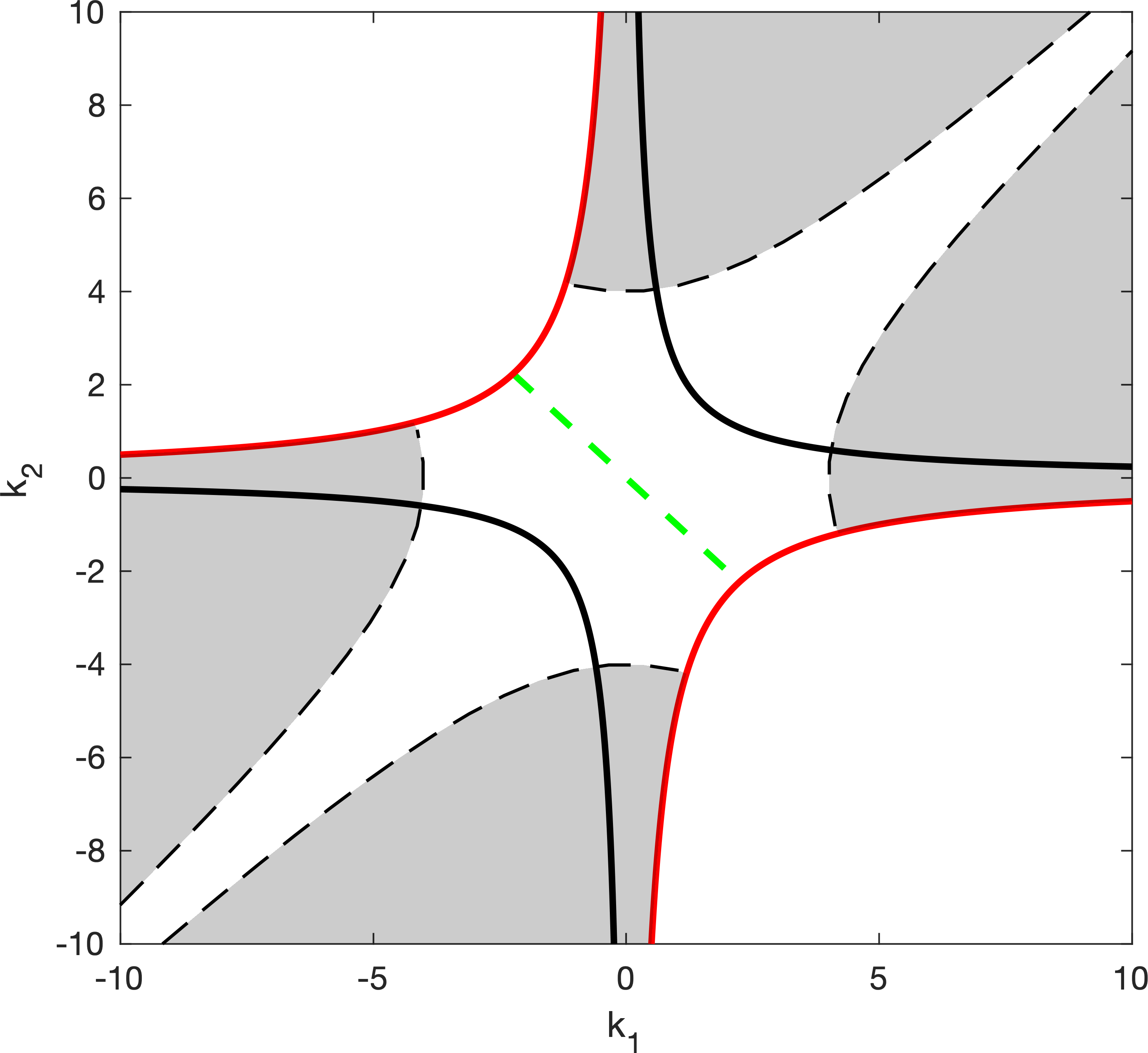}}%
\minp{0.57}{%
Example of phase plane for $\kappa_1\!=\!1$. 

\blue{Blue arrows:} vector field.

Black hyperbolas: equilibria $f(k_2k_1)=0$.

\red{red hyperbolas:} boundary of domain (stabilizing region).

%

\gray{Gray shaded area:}~region where imbalance $\geq\mu>0$

The imbalance measure is $c = \sqrt{\abs{{\KK}_1^2 \!-\! {\KK}_2^2}}$.

\green{Green dashed line:} stable manifold of saddle,  $k_1=-k_2^\top$.
}

\newsl{An ISS asymptotic gain property}

With $n\!=\!m\!=\!1$, $N\!=\!2$, arbitrary $\kappa_1$, consider the
gradient system
\[
\dot {\KK}_i = - \nabla_{{\KK}_i}\loss(\K) \, + \, u_i\,.
\]
Suppose that initial conditions satisfy
\[
\norm{k_1(0)+k_2(0)^\top}_2 \;>\; 2 \sqrt{a_+}, \quad (a_+:=\max\{a,0\})\,.
\]
%
Then: for every $\varepsilon>0$, there exists $\delta>0$
%
such that
\[
\norm{u_1}_\infty+\norm{u_2}_\infty
\leq \delta
\;\implies\;
\limsup_{t\rightarrow\infty} \left[\loss(k_2(t)k_1(t))-\lossmin\right] \;\leq\; \varepsilon 
\]
%

\minp{0.35}{\pic{0.35}{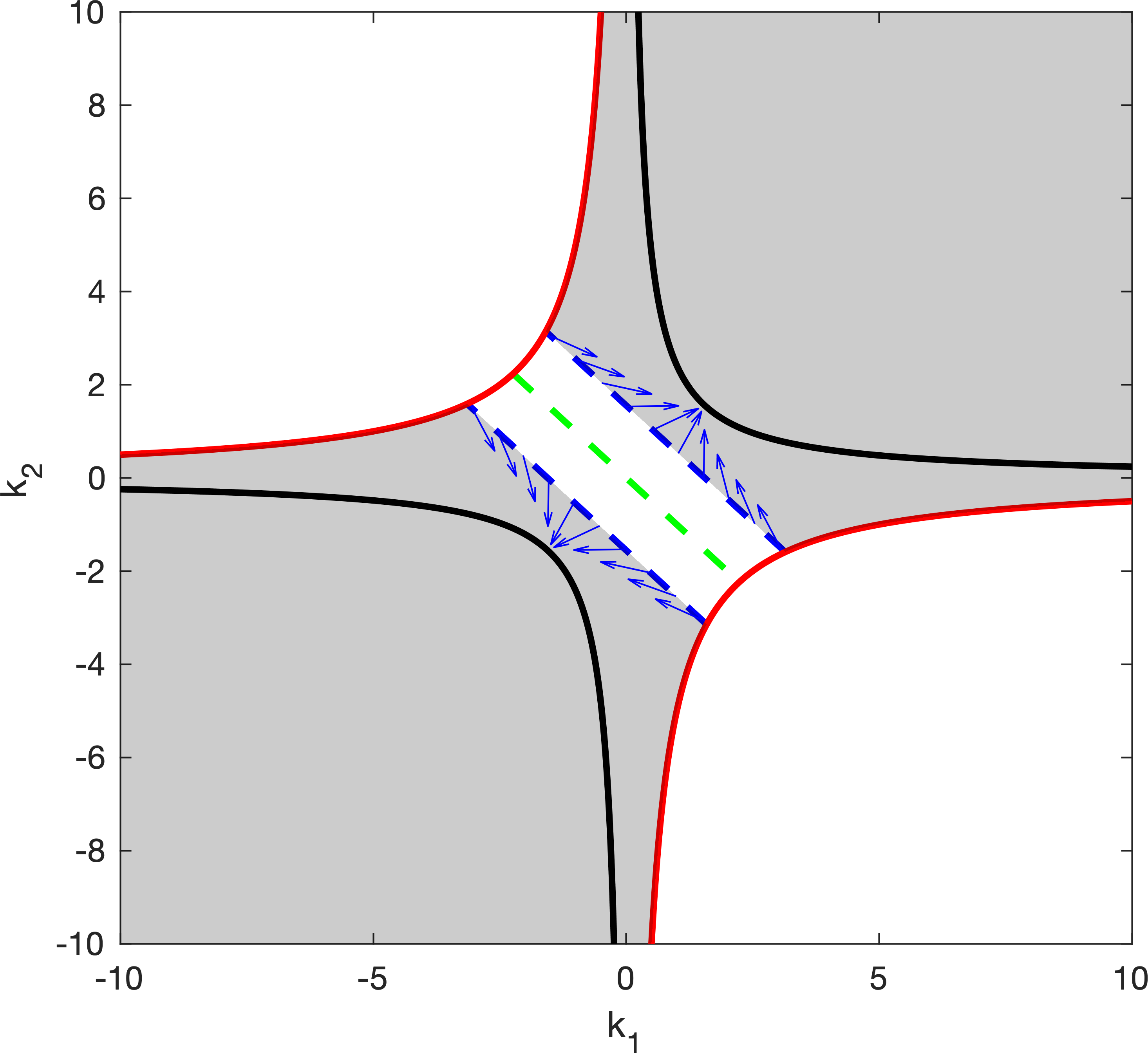}}%
\minp{0.65}{%
%
  %
Example of phase plane with $\kappa_1\!=\!1$ (and $a<0$).
  
\blue{Blue arrows:} vector field.

Black hyperbolas: equilibria $f(k_2k_1)=0$.

\red{Red hyperbolas:} boundary of domain (stabilizing region).

\blue{Blue segments:} $\norm{k_1 + k_2^\top}  = \alpha < 2\sqrt{k^*}$
[$k^* = a+\sqrt{a^2+q/r}$.] 
Note the transversality when crossing the blue segments, which leads
to robustness.

\gray{Gray shaded region:}~invariant region with (si)ISS stability.

\green{Green dashed line:} stable manifold of saddle,
$k_1=-k_2^\top$.
}

The theory has been worked out with $n\!=\!m\!=\!1$, but simulations
provide evidence for the advantage of higher imbalance for larger
dimensions.
Here is an example with here $n\!=\!5, m\!=\!3, N\!=\!2,
\kappa\!=\!10$.
The color code is:
darker blue means initial imbalance; red is classical solution.

\minp{0.5}{\pic{0.5}{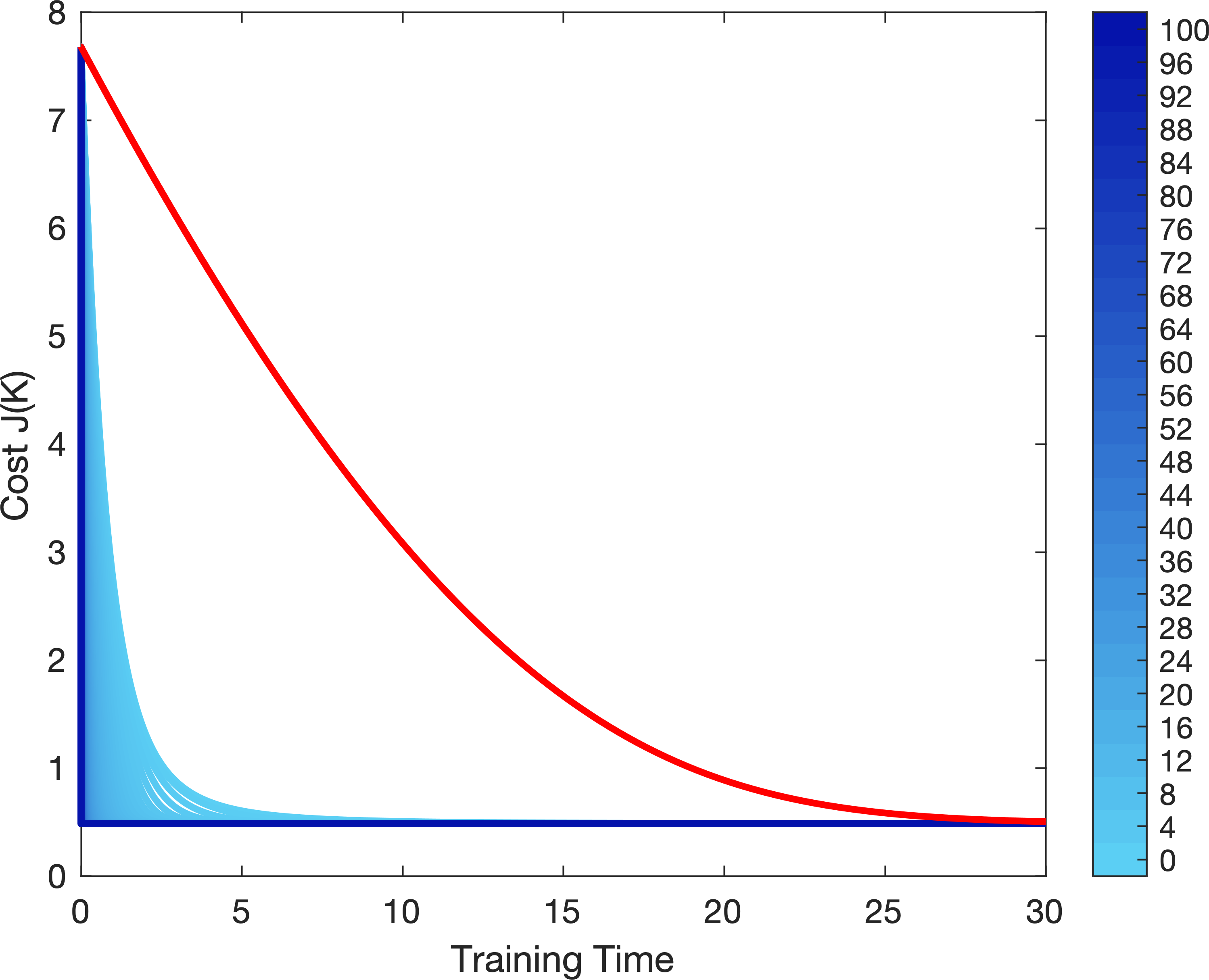}}%
\minp{0.5}{\pic{0.5}{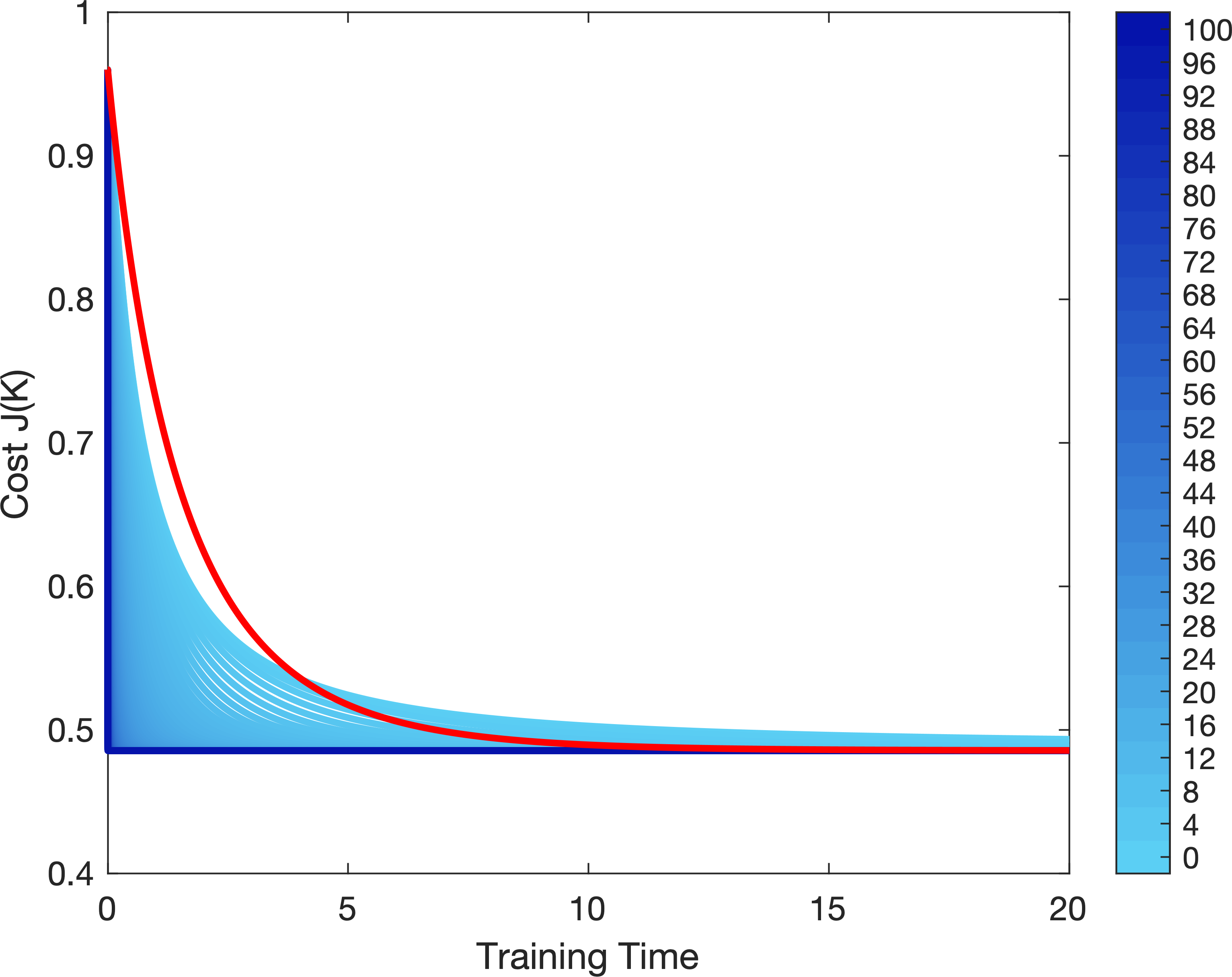}}

\ms

\minp{0.53}{{Classical (red): ``linear/exponential''.

Overparametrized: clearly exponential.

Larger imbalance$\implies$faster convergence.

[Even with no imbalance, still better!]}
}%
\minp{0.47}{{Slow-down near saddle.

Classical solution does not suffer from this.

But still faster initial convergence.}}

\section{Concluding remarks}

We saw that the classical $\gPLI$ gradient dominance condition can be
relaxed to several variants that are valid in interesting problems (for
example, $\satPLI$ for CT LQR problems).  Equally or more importantly,
these lead naturally to elegant characterizations of input to state
stability for disturbed gradient flows.

Furthermore, in the last section we saw how overparametrization may
allow one to recover $\gPLI$ on large invariant sets, and we
illustrated this with the LQR problem under (linear) neural network feedback.

Stay tuned for further results -- and in the meantime please consult the
published papers and preprints!

\newpage


\bibliographystyle{abbrvnat}
\bibliography{dynamics_learning}

\edo